\documentclass[sigconf, screen, authorversion]{acmart}
\usepackage{amsmath,amsfonts}
\usepackage{graphicx}
\usepackage[caption=false]{subfig}
\usepackage[ruled,linesnumbered,vlined]{algorithm2e}

\AtBeginDocument{%
  \providecommand\BibTeX{{%
    \normalfont B\kern-0.5em{\scshape i\kern-0.25em b}\kern-0.8em\TeX}}}

\setcopyright{acmlicensed}
\copyrightyear{2019}
\acmYear{2019}
\acmDOI{10.1145/1122445.1122456}

\acmConference[Anchorage '19]{Anchorage '19: ACM SIGKDD Workshop on  
Explainable AI/ML (XAI) for Accountability, Fairness, and Transparency}{August 04--08, 2019}{Anchorage, AK}
\begin{document}

%\settopmatter{printacmref=false} % Removes citation information below abstract
%\renewcommand\footnotetextcopyrightpermission[1]{} % removes footnote with conference information in first column
%\pagestyle{plain} % removes running headers

\makeatletter
\renewcommand\@formatdoi[1]{\ignorespaces}
\makeatother

%%
%% The "title" command has an optional parameter,
%% allowing the author to define a "short title" to be used in page headers.
\title[DLIME: A Deterministic Local Interpretable Model-Agnostic
Explanations]{DLIME: A Deterministic Local Interpretable Model-Agnostic Explanations Approach for Computer-Aided Diagnosis Systems}

%%
%% The "author" command and its associated commands are used to define
%% the authors and their affiliations.
%% Of note is the shared affiliation of the first two authors, and the
%% "authornote" and "authornotemark" commands
%% used to denote shared contribution to the research.
\author{Muhammad Rehman Zafar}
\affiliation{%
  \institution{Ryerson University}
  \city{Toronto}
  \country{Canada}}
\email{muhammadrehman.zafar@ryerson.ca}

\author{Naimul Mefraz Khan}
\affiliation{%
  \institution{Ryerson University}
  \city{Toronto}
  \country{Canada}}
\email{n77khan@ryerson.ca}

\begin{abstract}
Local Interpretable Model-Agnostic Explanations (LIME) is a popular technique used to increase the interpretability and explainability of black box Machine Learning (ML) algorithms. LIME typically generates an explanation for a single prediction by any ML model by learning a simpler interpretable model (e.g. linear classifier) around the prediction through generating simulated data around the instance by random perturbation, and obtaining feature importance through applying some form of feature selection. While LIME and similar local algorithms have gained popularity due to their simplicity, the random perturbation and feature selection methods result in \textit{instability} in the generated explanations, where for the same prediction, different explanations can be generated. This is a critical issue that can prevent deployment of LIME in a Computer-Aided Diagnosis (CAD) system, where stability is of utmost importance to earn the trust of medical professionals. In this paper, we propose a deterministic version of LIME. Instead of random perturbation, we utilize agglomerative Hierarchical Clustering (HC) to group the training data together and K-Nearest Neighbour (KNN) to select the relevant cluster of the new instance that is being explained. After finding the relevant cluster, a linear model is trained over the selected cluster to generate the explanations. Experimental results on three different medical datasets show the superiority for Deterministic Local Interpretable Model-Agnostic Explanations (DLIME), where we quantitatively determine the stability of DLIME compared to LIME utilizing the Jaccard similarity among multiple generated explanations.   
\end{abstract}


\maketitle

\section{Introduction}
\label{sec:intro}
AI and ML has has become a key element in medical imaging and precision medicine over the past few decades. However, most widely used ML models are opaque. In the health sector, sometimes the binary ``yes'' or ``no'' answer is not sufficient and questions like ``how'' or ``where'' something occurred is more significant. To achieve transparency, quite a few interpretable and explainable models have been proposed in recent literature. These approaches can be grouped based on different criterion \cite{molnar2019,guidotti2018assessing,plumb2018model} such as i) Model agnostic or model specific ii) Local, global or example based iii) Intrinsic or post-hoc iv) Perturbation or saliency based.
  
Among them, model agnostic approaches are quite popular in practice, where the target is to design a separate algorithm that can explain the decision making process of any ML model. LIME \cite{ribeiro2016should} is a well-known model agnostic algorithm. LIME is an instance-based explainer, which generates simulated data points around an instance through random perturbation, and provides explanations by fitting a sparse linear model over predicted responses from the perturbed points. The explanations of LIME are locally faithful to an instance regardless of classifier type. The process of perturbing the points randomly makes LIME a non-deterministic approach, lacking ``stability'', a desirable property for an interpretable model, especially in CAD systems.

In this paper, a Deterministic Local Interpretable Model-Agnostic Explanations (DLIME) framework is proposed. DLIME uses Hierarchical Clustering (HC) to partition the dataset into different groups instead of randomly perturbing the data points around the instance.  The adoption of HC is based on its deterministic characteristic and simplicity of implementation. Also, HC does not require prior knowledge of clusters and its output is a hierarchy, which is more useful than the unstructured set of clusters returned by flat clustering such as K-means \cite{manning2008iir}. Once the cluster membership of each training data point is determined, for a new test instance, KNN classifier is used to find the closest similar data points. After that, all the data points belonging to the predominant cluster is used to train a linear regression model to generate the explanations. Utilizing HC and KNN to generate the explanations instead of random perturbation results in consistent explanations for the same instance, which is not the case for LIME. We demonstrate this behavior through experiments on three benchmark datasets from the UCI repository, where we show both qualitatively and quantitatively how the explanations generated by DLIME are consistent and stable, as opposed to LIME.

\section{Related Work}
\label{sec:rwork}
For brevity, we restrict our literature review to locally interpretable models, which encourage understanding of learned relationship between input variable and target variable over small regions. Local interpretability is usually applied to justify the individual predictions made by a classifier for an instance by generating explanations. 

LIME \cite{ribeiro2016should} is one of the first locally interpretable models, which generates simulated data points around an instance through random perturbation, and provides explanations by fitting a sparse linear model over predicted responses from the perturbed points. In \cite{ribeiro2018anchors}, LIME was extended using decision rules. In the same vein, leave-one covariate-out (LOCO) \cite{lei2018distribution} is another popular technique for generating local explanation models that offer local variable importance measures. In \cite{hall2017machine}, authors proposed an approach to partition the dataset using K-means instead of perturbing the data points around an instance being explained. Authors in \cite{hu2018locally} proposed an approach to partition the dataset using a supervised tree based approach. Authors in \cite{katuwal2016machine} used LIME in precision medicine and discussed the importance of interpretablility to understand the contribution of important features in decision making.

Authors in \cite{Robnik2008} proposed a method to decompose the predictions of a classifier on individual contribution of each feature. This methodology is based on computing the difference between original predictions and predictions made by eliminating a set of features. In \cite{NIPS2017_7062}, authors have demonstrated the equivalence among various local interpretable models \cite{ijcai2017-371,NIPS2017_7272,Fong_2017_ICCV} and also introduced a game theory based approach to explain the model named SHAP (SHapley Additive exPlanations). 
Baehrens et al. \cite{baehrens2010explain} proposed an approach to yield local explanations using the local gradients that depict the movement of data points to change its expected label. A similar approach was used in \cite{Simonyan2013DeepIC,Zeiler2014,Zhou_2016_CVPR,Sundararajan2017} to explain and understand the behaviour of image classification models. 

One of the issues of the existing locally interpretable models is lack of ``stability''. In \cite{gosiewska2019ibreakdown}, this is defined as ``explanation level uncertainty'', where the authors show that explanations generated by different locally interpretable models have an amount of uncertainty associated with it due to the simplification of the black box model. In this paper, we address this issue at a more granular level. The basic question that we want to answer is: \textit{can explanations generated by a locally interpretable model provide consistent results for the same instance? } As we will see in the experimental results section, due to the random nature of perturbation in LIME, for the same instance, the generated explanations can be different, with different selected features and feature weights. This can reduce the healthcare practitioner's trust in the ML model. Hence, our target is to increase the stability of the interpretable model. Stability in our work specifically refers to  \textit{intensional stability} of feature selection method, which can be measured by the variability in the set of features selected \cite{Sarah2016,Kalousis2007}. Measures of stability include average Jaccard similarity and Pearson's correlation among all pairs of feature subsets selected from different training sets generated using cross validation, jacknife or bootstrap.

\section{Methodology}
\label{sec:method}

\begin{figure}[ht]
	\centering
		\includegraphics[width=\linewidth]{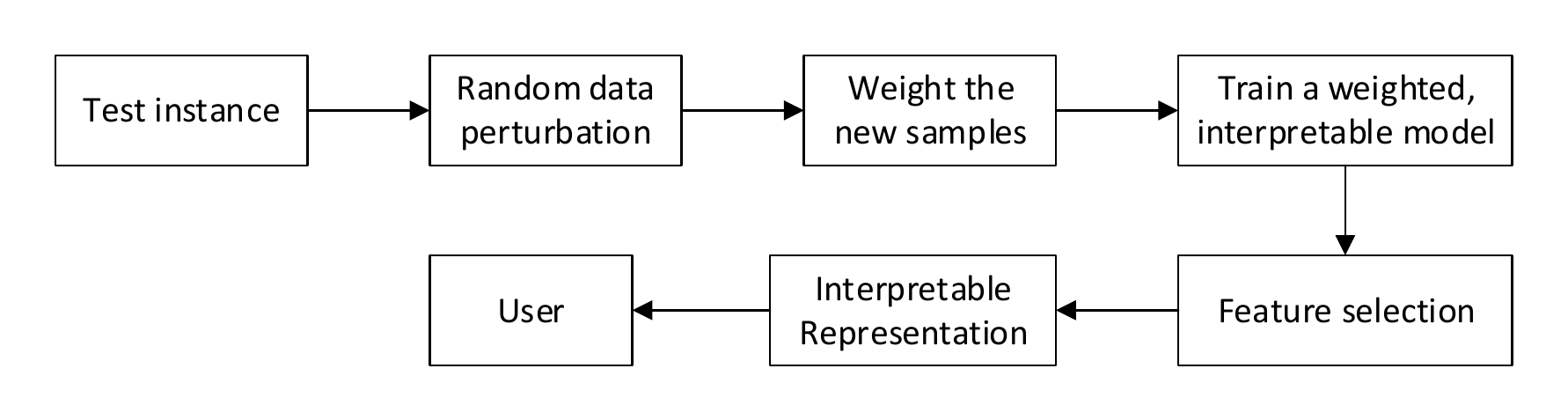}
	\caption{A block diagram of the LIME framework}
	\label{fig:lime}
\end{figure}
Before explaining DLIME, we briefly describe the LIME framework. LIME is a surrogate model that is used to explain the predictions of an opaque model individually. The objective of LIME is to train surrogate models locally and explain individual prediction.  Figure~\ref{fig:lime} shows a high level block diagram of LIME. It generates a synthetic dataset by randomly permuting the samples around an instance from a normal distribution, and gathers corresponding predictions using the opaque model to be explained. Then, on this perturbed dataset, LIME trains an interpretable model e.g. linear regression. Linear regression maintains relationships amongst variables which are dependent such as $Y$ and multiple independent attributes such as $X$ by utilizing a regression line $Y = a + bX$, where ``a'' is intercept, ``b'' is slope of the line. This equation can be used to predict the value of target variable from given predictor variables. In addition to that, LIME takes as an input the number of important features to be used to generate the explanation, denoted by $\mathcal{K}$. The lower the value of $\mathcal{K}$, the easier it is to understand the model. There are several approaches to select the  $\mathcal{K}$ important features such as i) backward or forward selection of features and ii) highest weights of linear regression coefficients. LIME uses the forward feature selection method for small datasets which have less than $6$ attributes, and highest weights approach for higher dimensional datasets. 

As discussed before, a big problem with LIME is the ``instability'' of generated explanations due to the random sampling process. Because of the randomness, the outcome of LIME is different when the sampling process is repeated multiple times, as shown in experiments. The process of perturbing the points randomly makes LIME a non-deterministic approach, lacking ``stability'', a desirable property for an interpretable model, especially in CAD systems.

\subsection{DLIME}
\begin{figure}[ht]
	\centering
		\includegraphics[width=\linewidth]{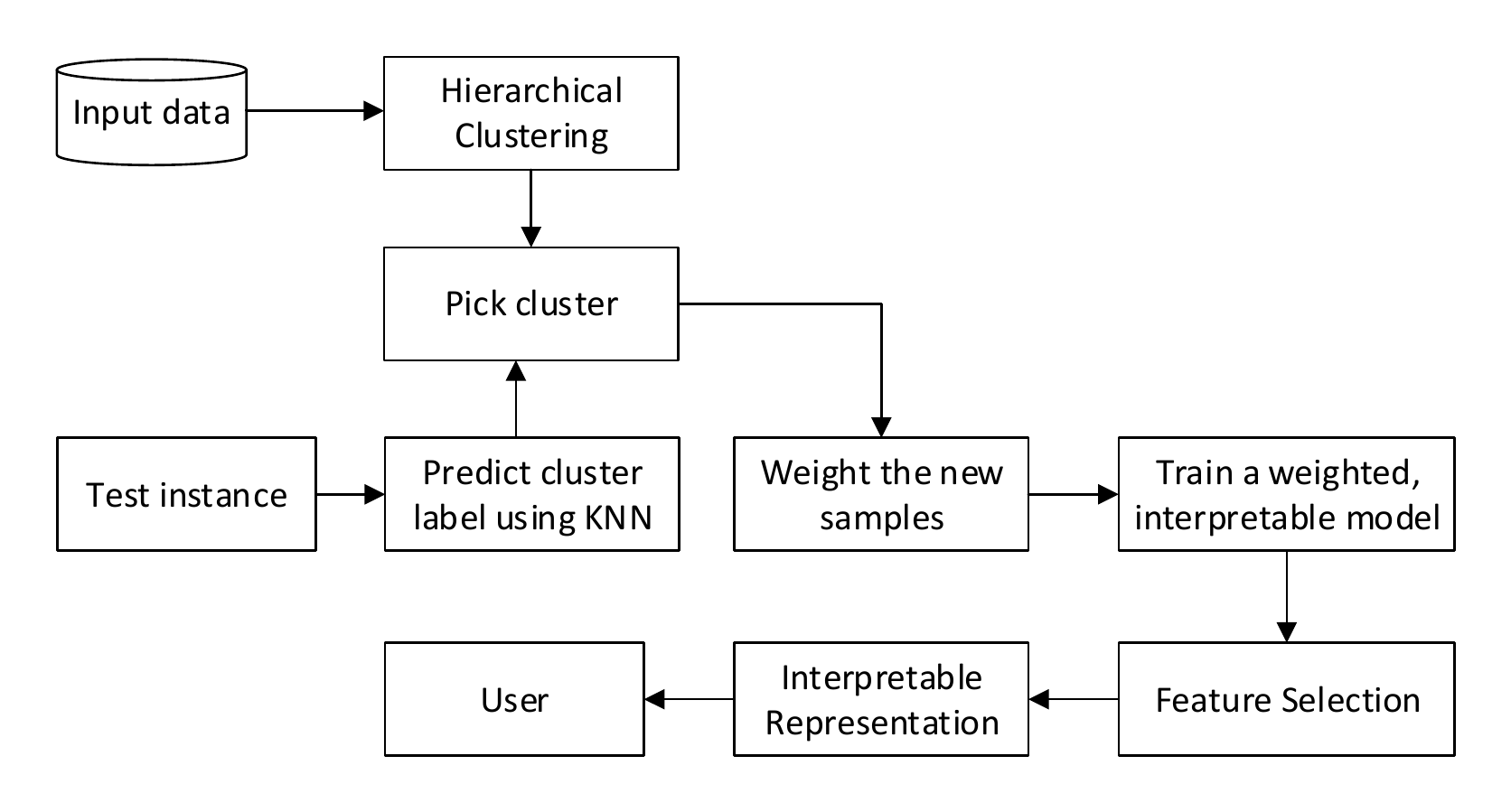}
	\caption{A block diagram of the DLIME framework}
	\label{fig:dlime}
\end{figure}

In this section, we present Deterministic Local Interpretable Model-Agnostic Explanations (DLIME), where the target is to generate consistent explanations for a test instance. Figure~\ref{fig:dlime} shows the block diagram of DLIME. 

The key idea behind DLIME is to utilize HC to partition the training dataset into different clusters. Then to generate a set of samples and corresponding predictions (similar to LIME), instead of random perturbation, KNN is first used to find the closest neighbors to the test instance. The samples with the majority cluster label among the closest neighbors are used as the set of samples to train the linear regression model that can generate the explanations. The different components of the proposed method are explained further below.

\subsubsection{Hierarchical Clustering (HC)}

HC is the most widely used unsupervised ML approach, which generates a binary tree with cluster memberships from a set of data points. The leaves of the tree represents data points and nodes represents nested clusters of different sizes. There are two main approaches to hierarchical clustering: divisive and agglomerative clustering. Agglomerative clustering follows the bottom-up approach and divisive clustering uses the top-down approach to merge the similar clusters. This study uses the traditional agglomerative approach for HC as discussed in \cite{duda1973pattern}. Initially it considers every data point as a cluster i.e. it starts with $N$ clusters and merge the most similar groups iteratively until all groups belong to one cluster. HC uses euclidean distance between closest data points or clusters mean  to compute the similarity or dissimilarity of neighbouring clusters \cite{heller2005bayesian}. 

\begin{figure}[h]
	\centering
		\includegraphics[scale=0.5]{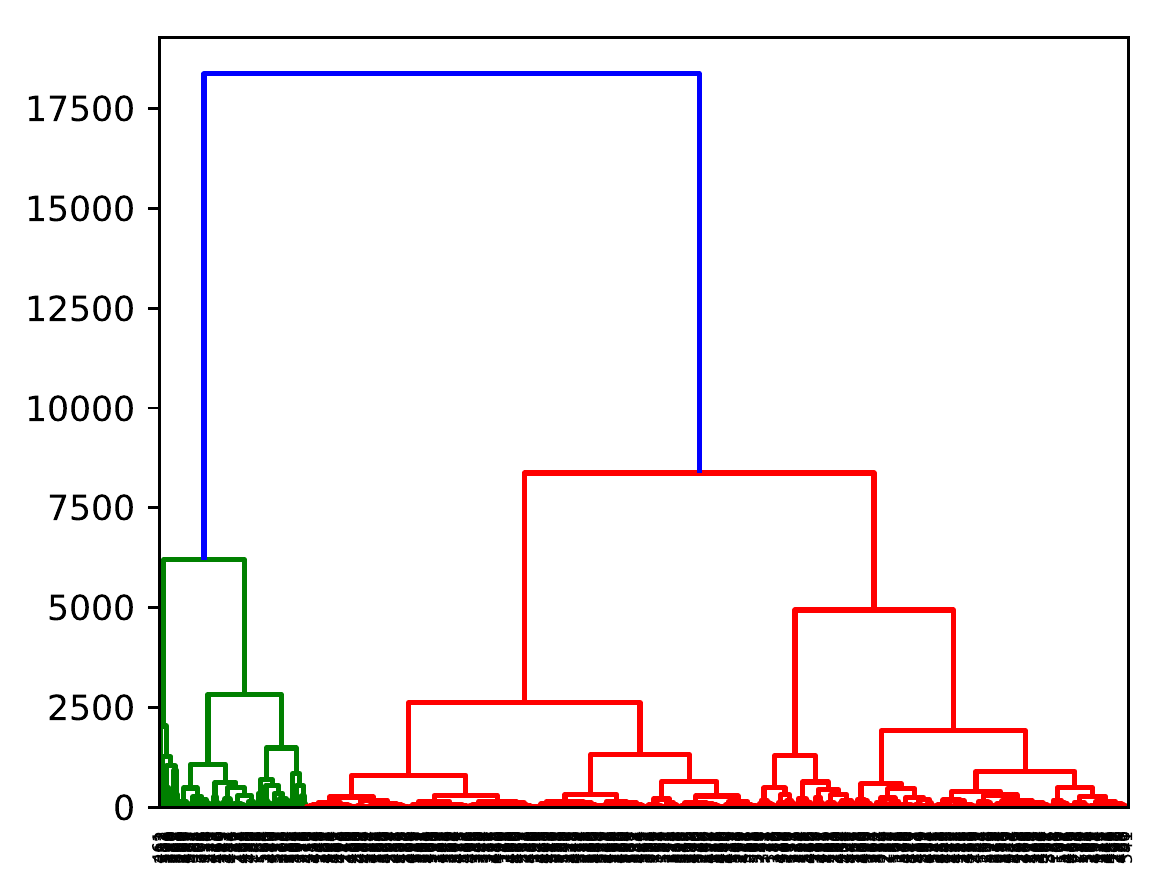}
	\caption{Dendrogram of breast cancer dataset}
	\label{fig:dend}
\end{figure}

One important step to use HC for DLIME is determining the appropriate number of clusters $C$. HC is in general visualized as a dendrogram. In dendrogram each horizontal line represents a merge and the y-coordinate of it is the similarity of the two merged clusters. A vertical line shows the gap between two successive clusters. By cutting the dendrogram where the gap is the largest between two successive groups, we can determine the value of $C$. As we can see in Figure~\ref{fig:dend} the largest gap between two clusters is at level 2 for the breast cancer dataset. Since all the datasets we have used for the experiments are binary, $C=2$ was always the choice from the dendrogram. For multi-class problems, the value of $C$ may change.

\subsubsection{K-Nearest Neighbor (KNN)}

Similar to other clustering approaches, HC does not assign labels to new instances. Therefore, KNN is trained over the training dataset to find the indices of the neighbours and predict the label of new instance. KNN is a simple classification model based on Euclidean distance \cite{cover1967knn}. It computes the distance between training and test sets. Let $x_i$ be an instance that belongs to a training dataset $\mathcal{D}_{train}$ of size $n$ where, $i$ in range of $1,2,... ,n$. Each  instance $x_i$ has $m$ features ($x_{i1}$,$x_{i2},... ,x_{im}$). In KNN, the Euclidean distance between new instance $x$ and a training instance $x_i$ is computed. After computing the distance, indices of the $k$ smallest distances are called $k$ -nearest neighbors. Finally, the cluster label for the test instance is assigned the majority label among the $k$-nearest neighbors. After that, all the data points belonging to the predominant class is used to train a linear regression model to generate the explanations. In our experiments, $k=1$ was used, as higher values of $k$ did not make a difference in the results.

Utilizing the notations introduced above, Algorithm~\ref{algo:hc} formally presents the proposed DLIME framework. 
 
\begin{algorithm}
\caption{Deterministic Local Interpretable Model-Agnostic Explanations}
\label{algo:hc}
\SetAlgoLined
\KwIn{Dataset $\mathcal{D}_{train}$, Instance $x$, length of explanation $\mathcal{K}$}
 Initialize $\mathcal{Y} \leftarrow \left\{\right\}$ \\
 Initialize clusters \For{i in 1 $\hdots$ N } { $C_i \leftarrow \left \{ i \right \}$ \\}
 Initialize clusters to merge $\mathcal{S} \leftarrow$ for i in 1 $\hdots$ N \\
\While{no more clusters are available for merging}
{
Pick two most similar clusters with minimum distance $d$: $(j,k)\leftarrow argmin_{d(j,k)} \in \mathcal{S}$ \\
Create new cluster $C_l \leftarrow C_j \cup C_k$ \\
Mark $j$ and $k$ unavailable to merge \\
\If{$C_l \neq i$ in $1 \hdots N $}{Mark $l$ as available, $\mathcal{S} \leftarrow \mathcal{S} \cup \left \{l \right \}$}
\ForEach{$i \in \mathcal{S}$}{Update similarity matrix by computing distance $d(i,l)$}
}
 %\KwRet$C$  \\
 %Get labels and derive a new feature called "membership" in $\mathcal{D}_{train}$\\
\While{i in 1, ... , n}
{
$d({\textbf x}_i,{\textbf x}) = \sqrt{(x_{i1}-x_{1})^2 +  \cdots + (x_{im}-x_{m})^2}$
}
$ind \leftarrow $ Find indices for the $k$ smallest distance $d({\textbf x}_i,{\textbf x})$ \\
$\hat{y} \leftarrow$ Get majority label for $x \in ind$ \\
$n^{s} \leftarrow$ Filter $\mathcal{D}_{train}$ based on $\hat{y}$ \\
% \KwRet $n^{s}$ 
\ForEach{i in 1,... ,n}
 {
 %$n^s_i \leftarrow $ find\_cluster($x$) using algorithm~\ref{algo:knn} \\
 %$\mathcal{Y} \leftarrow \mathcal{Y} \cup (b,m,\pi_x(n^{s}_{i}))$
 $\mathcal{Y} \leftarrow  $ Pairwise distance of each instance in cluster $n^{s}$ with the original instance $x$
 }
 %$\omega \leftarrow$ K-Lasso$(\mathcal{Y},\mathcal{K})$  \\
 $\omega \leftarrow$ LinearRegression$(n^{s}, \mathcal{Y},\mathcal{K})$  \\
 %Filter $\mathcal{K}$ explanations from $\mathcal{Y}$ interpretable space ($\mathcal{Y}, \mathcal{K}$) \\
 \KwRet $\omega$ 
\end{algorithm}

\section{Experiments}
\subsection{Dataset} \label{sec:data}
To conduct the experiments we have used the following three healthcare datasets from the UCI repository \cite{Dua2019}.

\subsubsection{Breast Cancer} \label{bc}
The first case study is demonstrated by using the widely adopted Breast Cancer Wisconsin (Original) dataset. The dataset consists of $699$ observations and $11$ features \cite{mangasarian1995breast}.

\subsubsection{Liver Patients} \label{lp}
The second case study is demonstrated by using an Indian liver patient dataset. The dataset consists of $583$ observations and $11$ features \cite{ramana2011critical}. 

\subsubsection{Hepatitis Patients} \label{hp}
The third case study is demonstrated by using hepatitis patients dataset. The dataset consists of $20$ features and $155$ observations out of which only $80$ were used to conduct the experiment after removing the missing values \cite{diaconis1983computer}. 

\subsection{Opaque Models}
We trained two opaque models from the scikit-learn package \cite{scikit-learn} over the three datasets:  Random Forest (RF) and Neural Networks (NN), which are difficult to interpret without an explainable model.

\subsubsection{Random Forest}
Random Forest (RF) is a supervised machine learning algorithm that can be used for both regression and classification \cite{biau2016random}. RF constructs many decision trees and selects the one which best fits on the input data. The performance of RF is based on the correlation and strength of each tree. If the correlation between two trees is high the error will also be high. The tree with the lowest error rate is the strength of the classifier. 
\subsubsection{Neural Network}
Neural Network (NN) is a biologically inspired model, which is composed of a large number of highly interconnected neurons working concurrently to resolve particular problems. There are different types of NNs, among which we picked the popular feed forward artificial NN. It typically has multiple layers and is trained with the backpropagation. The particular NN we utilized has two hidden layers. The first hidden layer has $5$ hidden units and the second hidden layer has 2 hidden units.

For both opaque models, $80\%$ data is used for training and the remaining $20\%$ data is used for evaluation. Here, it is worth mentioning that, both NN and RF models scored over 90\% accuracy on each dataset which is reasonable. Their performance can be improved by hyper parameter tuning. However, our aim is to produce deterministic explanations, therefore we have not spent additional effort to further tune these models.

After training these opaque models, both LIME (default python implementation) and DLIME were used to generate explanations for a randomly selected test instance. For the same instance, 10 iterations of both algorithms were executed to determine stability. 

\subsection{Results}
\begin{figure*}
\centering
\subfloat[Iteration 1: Explanations generated with DLIME for NN]{\includegraphics[clip,width=0.5\linewidth]{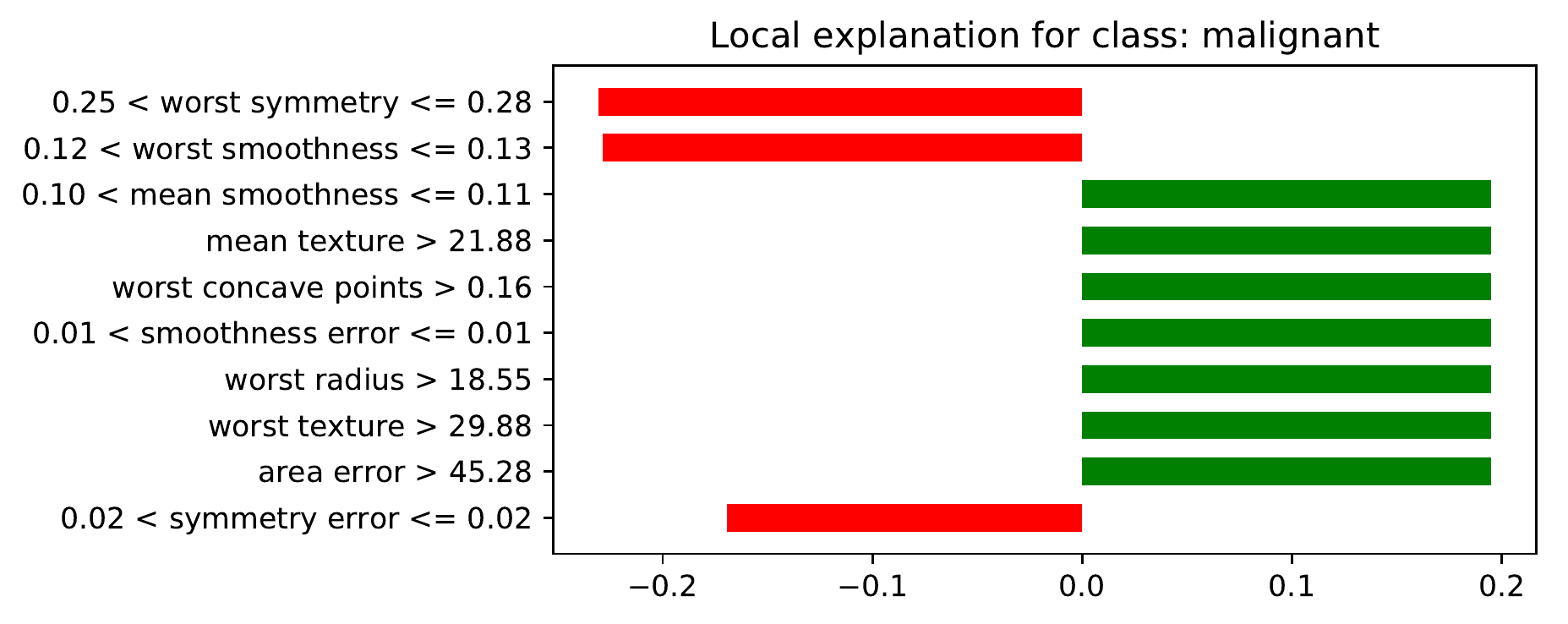}}
\subfloat[Iteration 1: Explanations generated with LIME for NN]{\includegraphics[clip,width=0.5\linewidth]{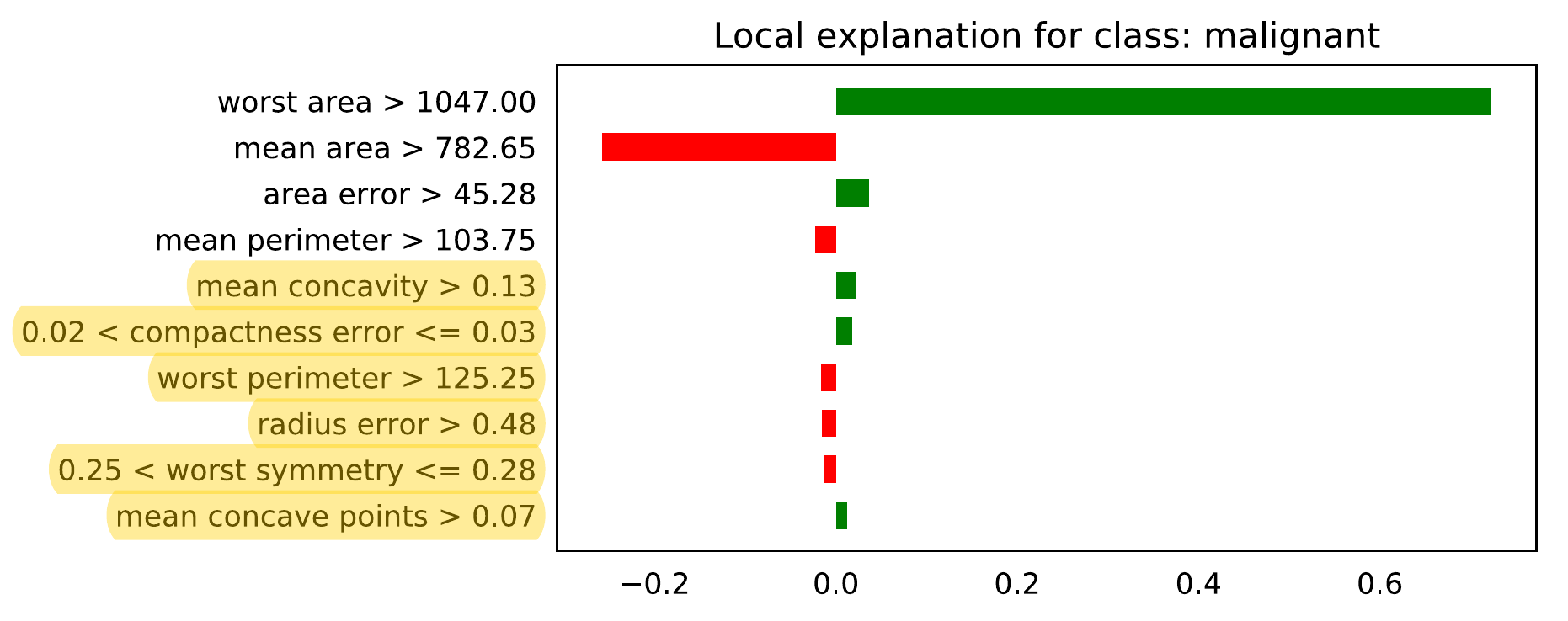}}

\subfloat[Iteration 2: Explanations generated with DLIME for NN]{\includegraphics[clip,width=0.5\linewidth]{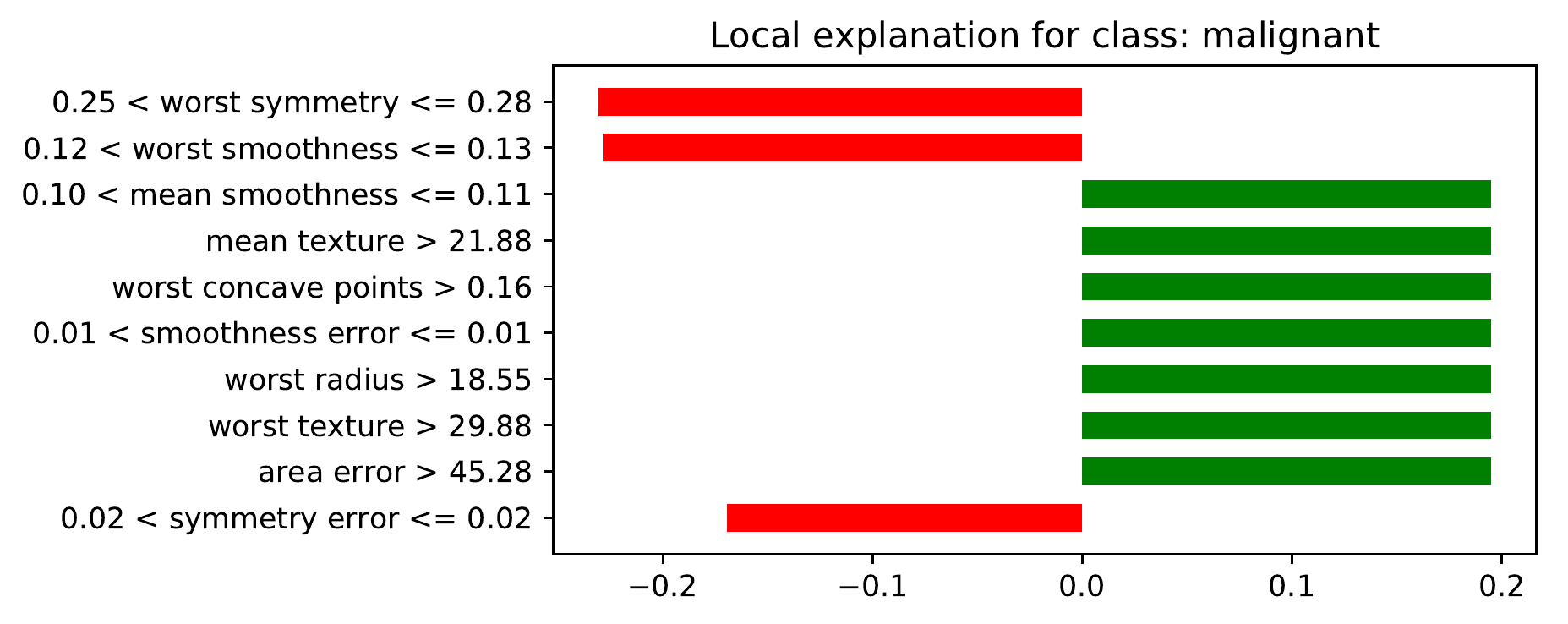}}
\subfloat[Iteration 2: Explanations generated with LIME for NN]{\includegraphics[clip,width=0.5\linewidth]{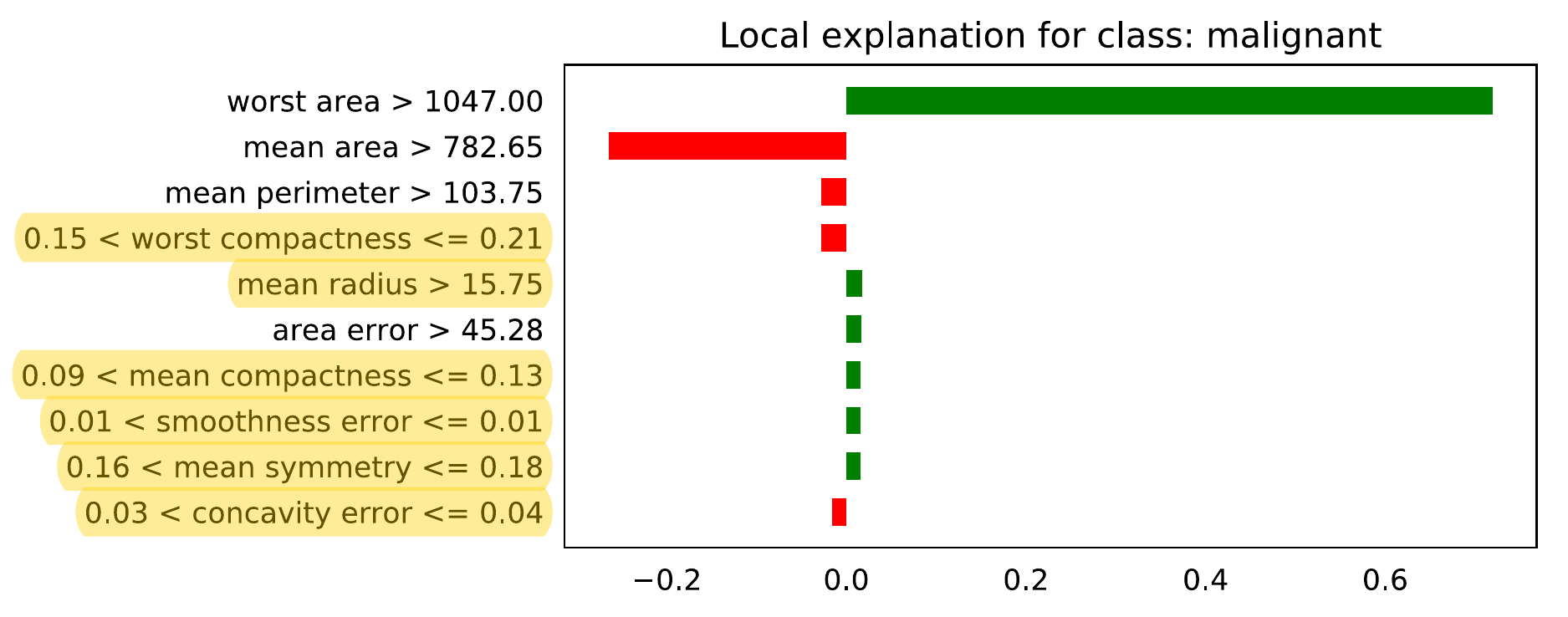}}

\subfloat[Jaccard distance of DLIME explanations]{\includegraphics[clip,scale=0.5]{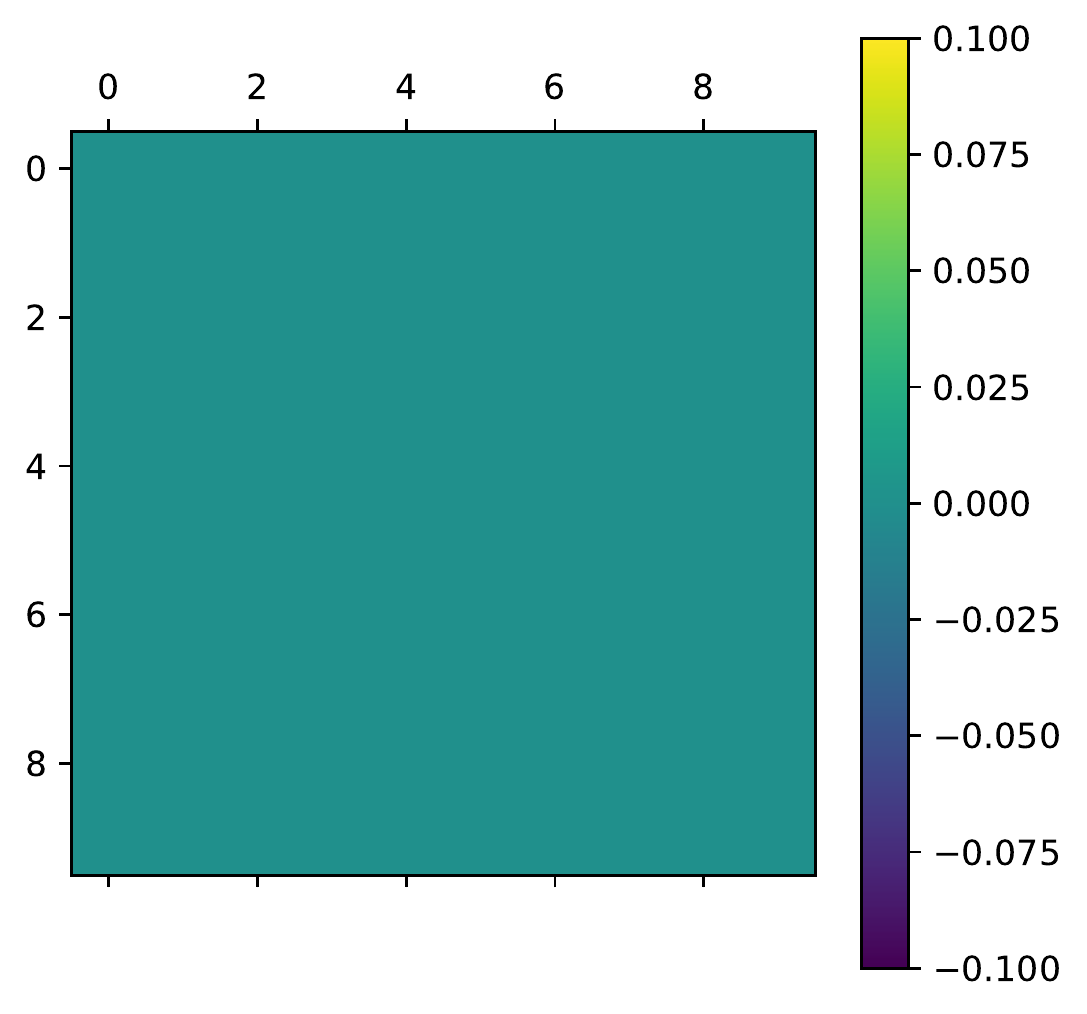}}
\subfloat[Jaccard distance of LIME explanations]{\includegraphics[clip,scale=0.5]{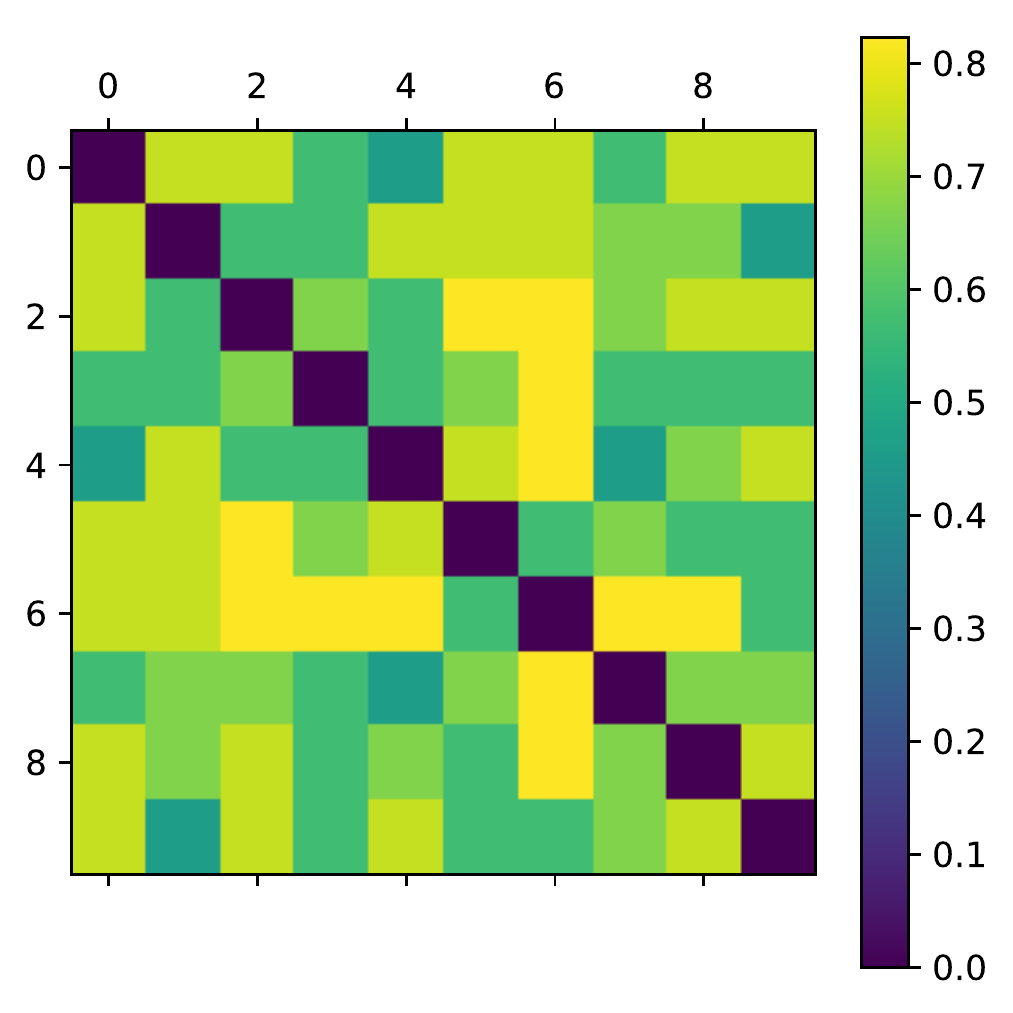}}

\caption{Explanations generated by DLIME and LIME, and respective Jaccard distances over 10 iterations}
\label{fig:bcucnn}
\end{figure*}

Fig.~\ref{fig:bcucnn} shows the results for two iterations of explanations generated by DLIME and LIME for a randomly selected test instance with the trained NN on  the breast cancer dataset. On the left hand side in  Fig.~\ref{fig:bcucnn} (a) and Fig.~\ref{fig:bcucnn} (c) are the explanations generated by DLIME, and on the right hand side Fig.~\ref{fig:bcucnn} (b) and Fig.~\ref{fig:bcucnn} (d) are the explanations generated by LIME. The red bars in Fig.~\ref{fig:bcucnn} (a), shows the negative coefficients and green bars shows the positive coefficients of the linear regression model. The positive coefficients shows the positive correlation among the dependent and independent attributes. On the other hand negative coefficients shows the negative correlation among the dependent and independent attributes. 

As we can see, LIME is producing different explanations for the same test instance. The yellow highlighted attributes in Fig.~\ref{fig:bcucnn} (b) are different from those in Fig.~\ref{fig:bcucnn} (d).  On the other hand, explanations which are generated with DLIME are deterministic and stable as shown in Fig.~\ref{fig:bcucnn} (a) and Fig.~\ref{fig:bcucnn} (b). Here, it is worth mentioning that, both DLIME and LIME frameworks may generate different explanations. LIME is using $5000$ randomly perturbed data points around an instance to generate the explanations. On the other hand, DLIME is using only clusters from the original dataset. For DLIME linear regression, the dataset has less number of data points since it is limited by the size of the cluster. Therefore, the explanations generated with LIME and DLIME can be different. However, DLIME explanations are stable, while those generated by LIME are not. 

To further quantify the stability of the explanations, we have used the Jaccard coefficient. The Jaccard coefficient is a similarity and diversity measure among finite sets. It computes the similarity between two sets of data points by computing the number of elements in intersection divided by the number of elements in union. The mathematical notation is given in equation~\ref{eq:jaccard}.

\begin{equation}
\label{eq:jaccard}
J(S_1, S_2) = \frac{|S_1 \cap S_2|}{|S_1 \cup S_2|}
\end{equation}

Where, $S_1$ and $S_2$ are the two sets of explanations. The result of $J(S_1, S_2) = 1$ means $S_1$ and $S_2$ are highly similar sets. $J(S_1, S_2) = 0$ when $|S_1 \cap S_2| = 0$, implying that $S_1$ and $S_2$ are highly dissimilar sets. 

Based on the Jaccard coefficient, Jaccard distance is defined in equation~\ref{eq:jdist}.
\begin{equation}
\label{eq:jdist}
J_{distance} = 1- J(S_1, S_2)
\end{equation} 

To quantify the stability of DLIME and LIME, after 10 iterations, $J_{distance}$ is computed among the generated explanations. 

Fig.~\ref{fig:bcucnn} (e) and Fig.~\ref{fig:bcucnn} (f) shows the $J_{distance}$. It is a $10 \times 10$ matrix. The diagonal of this matrix is $0$ that shows the $J_{distance}$ of the explanation with itself and lower and upper diagonal shows the $J_{distance}$ of explanations from each other. Lower and upper diagonal are representing the same information. It can be observed that, the $J_{distance}$ in Fig.~\ref{fig:bcucnn} (e) is $0$ which means the generated explanations with DLIME are deterministic and stable on each iteration. However, for LIME, as we can see, the $J_{distance}$ contain significant values, further proving the instability of LIME.

Finally, Table \ref{tab:jd} lists the average  $J_{distance}$ obtained for DLIME and LIME utilizing the two opaque models on the three datasets (10 iterations for one randomly selected test instance). As we can see, in every scenario, the $J_{distance}$ for DLIME is zero, while for LIME it contains significant values, demonstration the stability of DLIME when compared with LIME. 

\begin{table}
  \caption{Average $J_{distance}$ after $10$ iterations}
  \label{tab:jd}
  \begin{tabular}{ |l|c|c|c| } 
  \hline
  Dataset & Opaque Model & DLIME & LIME \\
   \hline
   Breast Cancer & RF & 0 &9.43\% \\
   Breast Cancer & NN & 0 &57.95\% \\ 
   Liver Patients & RF & 0 &17.87\% \\
   Liver Patients & NN & 0 &55.00\% \\
   Hepatitis Patients & RF & 0 &16.46\% \\
   Hepatitis Patients & NN & 0 &39.04\% \\
   \hline
 \end{tabular}
\end{table}

\section{Conclusion}
\label{sec:conclude}

In this paper, we propose a deterministic approach to explain the decisions of black box models. Instead of random perturbation, DLIME uses HC to group the similar data in local regions, and utilizes KNN to find cluster of data points that are similar to a test instance. Therefore, it can produce deterministic explanations for a single instance. On the other hand, LIME and other similar model agnostic approaches based on random perturbing may keep changing their explanation on each iteration, creating distrust particularly in medical domain where consistency is highly required. We have evaluated DLIME on three healthcare datasets. The experiments clearly demonstrate that the explanations generated with DLIME are stable on each iteration, while LIME generates unstable explanations. 
 
Since DLIME depends on hierarchical clustering to find similar data points, the number of samples in a dataset may affect the quality of clusters, and consequently, the accuracy of the local predictions. In future, we plan to investigate how to solve this issue while keeping the model explanations stable. We also plan to experiment with other data types, such as images and text.

Keeping up with the sipirit of reproducible research, all datasets and source code can be accessed through the Github repository \url{https://github.com/rehmanzafar/dlime_experiments.git}.

%\vfill \pagebreak
%\bibliographystyle{splncs04-unsrt}
\bibliographystyle{ACM-Reference-Format}
%\balance 
%\bibliography{bib_reference_list}

\end{document}